\documentclass[11pt]{article}

\usepackage[final]{acl}
\usepackage{times}
\usepackage{latexsym}
\usepackage{drs}
\usepackage[T1]{fontenc}
\usepackage[utf8]{inputenc}
\usepackage{microtype}
\usepackage{inconsolata}
\usepackage{graphicx,xcolor}
\usepackage{url}
\usepackage{linguex}
\usepackage{booktabs}
\usepackage{listings}
\usepackage{amsmath}
\usepackage{tabularx}
\usepackage{cancel}
\usepackage{subcaption}
\usepackage{graphicx}

\newcommand{\todo}[1]{\textcolor{black}{#1}}
\usepackage{tabularx, booktabs, array}
\newcolumntype{Y}{>{\raggedright\arraybackslash}X}
\newcommand{\LF}[1]{\ensuremath{\texttt{#1}}}
\newcommand{\setof}[1]{\ensuremath{\left \{ #1 \right \}}}
\newcommand{\tuple}[1]{\ensuremath{\left \langle #1 \right \rangle }}

\setkeys{Gin}{width=\linewidth,height=\textheight,keepaspectratio}

\title{A Theorem-Proving-Based Evaluation of Neural Semantic Parsing}

\author{
 \textbf{Hayate Funakura\textsuperscript{1,2,3}},
 \textbf{Hyunsoo Kim\textsuperscript{2}},
 \textbf{Koji Mineshima\textsuperscript{2}}
\\
\\
 \textsuperscript{1}Kyoto University
 \textsuperscript{2}Keio University
 \textsuperscript{3}Kikagaku Inc.
\\
 \small{
   \textbf{Correspondence:} \href{mailto:funakura.hayate.28p@st.kyoto-u.ac.jp}{funakura.hayate.28p@st.kyoto-u.ac.jp}
 }
}

\begin{document}
\maketitle

\begin{abstract}
Graph-matching metrics such as Smatch are the de facto standard for evaluating neural semantic parsers, yet they capture surface overlap rather than logical equivalence. We reassess evaluation by pairing graph-matching with automated theorem proving. We compare two approaches to building parsers: supervised fine-tuning (T5-Small/Base) and few-shot in-context learning (GPT-4o/4.1/5), under normalized and unnormalized targets.
We evaluate outputs using graph-matching, bidirectional entailment between source and target formulas with a first-order logic theorem prover, and well-formedness. Across settings, we find that models performing well on graph-matching often fail to produce logically equivalent formulas.
Normalization reduces incidental target variability, improves well-formedness, and strengthens logical adequacy. Error analysis shows performance degrades with increasing formula complexity and with coordination, prepositional phrases, and passive voice; the dominant failures involve variable binding and indexing, and predicate naming. These findings highlight limits of graph-based metrics for reasoning-oriented applications and motivate logic-sensitive evaluation and training objectives together with simplified, normalized target representations. All code and data for our experiments are publicly available.\footnote{\url{https://github.com/hfunakura/text2sem}}
\end{abstract}

\section{Introduction}
\label{sec:intro}

Semantic parsing is the task of mapping natural language expressions into structured representations such as database queries or logical forms. These outputs have a wide range of applications, including document classification~\cite{dong2015statistical} and question answering~\cite{yih2014semantic}. Neural network-based approaches have become prominent in semantic parsing~\cite{konstas-etal-2017-neural, bai2022graph}, with Smatch \cite{cai2013smatch} widely used for evaluation. Smatch compares two Abstract Meaning Representations (AMRs)~\cite{banarescu-etal-2013-abstract}
by aligning their atomic propositions and computing the F-score of the overlap. We refer to Smatch and its variants as \textit{graph-matching-based} evaluation.

The aim of this paper is to reconsider evaluation methods for neural semantic parsing from the perspective of logical reasoning. 
One key use of the logical forms produced by semantic parsing is to support logically correct inference.
For such inference, the semantic parser must generate formal representations that enable a symbolic solver (e.g., an automated theorem prover) to derive correct outcomes such as entailment, contradiction, or consistency.
Ideally, for a sentence $S$ with a gold semantic representation $\mathrm{SR}_g(S)$, the parser's output $\mathrm{SR}_p(S)$ should be logically equivalent to $\mathrm{SR}_g(S)$.
However, graph-matching-based evaluation focuses solely on surface overlap between graph components, and thus may fail to reflect whether the predicted and gold representations are truly equivalent in meaning.

In this paper, we test the hypothesis that a model achieving high performance in graph-matching-based evaluation does not necessarily perform well in evaluation aimed at accurate natural language inference. We evaluate models that convert English sentences into first-order predicate logic representations using both evaluation methods. 
We consider two model settings: supervised fine-tuning (SFT) of a Transformer-based pre-trained semantic parser, and in-context learning (ICL) with several pre-trained language models of different sizes, including the latest GPT-5, where a few parsing examples are given before parsing unseen sentences.

Our contributions are threefold:

\paragraph{Limits of graph-based metrics} 
Pairing graph-matching with theorem-prover entailment reveals a gap between surface overlap and logical equivalence: models with high graph-matching scores often fail to produce logically correct predictions.

\paragraph{Benefit of target normalization}
Training on normalized formulas consistently improves performance by reducing incidental variability and enhancing well-formedness and logical adequacy.

\paragraph{Error patterns and next steps} 
Performance declines with formula complexity and with coordination, prepositional phrases, and passive voice. The most frequent errors involve variable binding and indexing and predicate naming, motivating stronger handling of linguistic phenomena and simpler target representations.

\section{Background and Related Work}
\label{sec:background}
\subsection{Semantic Parsing}
\label{ssec:parsing}

The task of converting natural language expressions into formal semantic representations has been extensively studied in the fields of symbolic logic and formal semantics~\cite{blackburn2005representation,lepore2009}.
Advances in this area have been accelerated by the development of syntactically expressive grammar formalisms such as Combinatory Categorial Grammar (CCG)~\cite{Steedman2000}, the creation of linguistically rich resources like CCGbank~\cite{hockenmaier2007ccgbank}, and the emergence of wide-coverage semantic parsers enabled by progress in syntactic parsing technologies~\cite{bos2004wide}.

More recently, the development of semantically annotated corpora such as 
AMR~\cite{banarescu-etal-2013-abstract} and Parallel Meaning Bank (PMB)~\cite{abzianidze-etal-2017-parallel} has accelerated research into neural approaches to semantic parsing. In particular, sequence-to-sequence models have become widely adopted for learning mappings from natural language to logical forms. These models have been successfully applied to a variety of downstream tasks, including code generation~\cite{ling-etal-2016-latent}, question answering~\cite{dong-lapata-2016-language}, and natural language generation~\cite{konstas-etal-2017-neural}.

Originally, semantic parsing was considered a promising approach for enabling and improving a wide range of downstream tasks requiring semantic understanding, including translation, summarization, question answering, and paraphrasing.
Graph-matching-based evaluation methods, such as Smatch~\cite{cai2013smatch} and subsequent variants for AMR~\cite{opitz-etal-2020-amr,opitz-2023-smatch}, as well as adaptations for Discourse Representation Structure (DRS)~\cite{van-noord-etal-2018-evaluating}, were developed with this broad applicability in mind and provide flexible means to assess parsing performance across diverse use cases.
However, with the rise of large pre-trained language models~\cite{devlin2019bert,brown2020language}, the role of formal semantic representations in these downstream tasks has lessened, and the relevance of semantic parsing itself has come under renewed scrutiny~\cite{van-noord-etal-2020-character}.

\subsection{Logical Entailment}
\label{ssec:entailment}

One core task where formal semantic representations play a role is recognizing logical entailment in natural language inference, typically formulated as determining whether a premise entails, contradicts, or is neutral with respect to a hypothesis.
Traditional logical formalisms developed within symbolic logic~\cite{blackburn2005representation} were often designed with the goal of enabling precise logical reasoning---such as entailment and consistency checking---by combining these representations with Automated Theorem Proving (ATP) techniques~\cite{fitting1996first,robinson2001handbook}. An early attempt to directly apply this paradigm to natural language entailment recognition was proposed by~\newcite{bos-markert-2005-recognising}. Further work has extended this line of research by leveraging semantic parsing based on CCG in combination with theorem provers to handle a wide range of natural language inferences~\cite{abzianidze:2015:EMNLP,mineshima-etal-2015-higher,Haruta_Mineshima_Bekki_2022}.
In contrast, the effectiveness of combining neural semantic parsing with automated theorem proving for natural language inference remains, to the best of our knowledge, an open question that has not yet been fully explored.

One of the domains where precise logical reasoning is essential is mathematical theorem proving. The task of converting natural language proofs into formal representations that can be handled by automated theorem provers or interactive proof assistants such as Coq and Lean has been extensively studied under the name of \emph{autoformalization}~\cite{wu2022autoformalization}.
Whether pre-trained language models alone can support the kind of rigorous logical reasoning required for complex problem solving remains an open question.
At present, there is ongoing exploration into hybrid approaches that integrate structured, logic-based semantic representations, which guarantee precision and correctness, with statistical language models~\cite{kautz2022third}.

\begin{table*}[ht]
\centering
\caption{Example sentences and their event-semantic representations. In the IDs, ``\texttt{p}'' indicates the premise and ``\texttt{h}'' indicates the hypothesis.
In the semantic representations, variables of the form $x_1, x_2, \ldots$ are used for entities, while $e_1, e_2, \ldots$ are used for events. The hyphen ``\LF{-}'' denotes logical (boolean) negation.
``Complexity'' counts the number of logical constants (negation, quantifier, and conjunction).
``\LF{exists e1 x2 x3}'' is an abbeviation for ``\LF{exists e1. exists e2. exists e3}''.
}
\begin{tabularx}{\textwidth}{lXc}
\hline
\textbf{ID} & \texttt{sick\_train\_88\_p} & Complexity \\
\hline
(a) Sentence & There is no biker jumping in the air. & ---\\
(b) Raw & \texttt{-exists\allowbreak{} x5.(\_biker(x5) \& exists e6.(\_jump(e6) \& (subj(e6) = x5) \& exists x7.(\_air(x7) \& \_in(e6,x7))))} & 8 \\
(c) Prenex & \texttt{-exists\allowbreak{} e1\allowbreak{} x2\allowbreak{} x3.(biker(x2) \& jump(e1) \& (subj(e1) = x2) \& air(x3) \& in(e1,x3))} & 8 \\
\hline
\textbf{ID} & \texttt{sick\_train\_55\_p} \\
\hline
(a) Sentence & Three boys are jumping in the leaves. & --- \\
(b) Raw & \texttt{exists\allowbreak{} x4.(\_boy(x4) \& \_three(x4) \& exists e5.(\_jump(e5) \& (subj(e5) = x4) \& exists x6.(\_leaf(x6) \& \_in(e5,x6))))} & 8 \\
(c) Prenex & \texttt{exists\allowbreak{} e1\allowbreak{} x2\allowbreak{} x3.(boy(x2) \& three(x2) \& jump(e1) \& (subj(e1) = x2) \& leaf(x3) \& in(e1,x3))} & 8 \\
\hline
\end{tabularx}
\label{tab:example-semantic}
\end{table*}

\subsection{Compositional Generalization}
\label{ssec:comp}
Another domain where neural semantic parsing has been actively studied is compositional generalization, which examines whether a model can generalize to novel syntactic and semantic combinations when mapping natural language expressions to logical forms. A widely used benchmark in this area is COGS~\cite{kim-linzen-2020-cogs}, followed by numerous extensions, most of which rely on exact matching of predicted and gold logical forms as the evaluation metric. However, exact matching treats logically equivalent expressions as different---for example, it fails to recognize the equivalence of conjunctive forms such as $p \wedge q$ and $q \wedge p$, or of formulas that differ only in variable naming, such as $\exists x_1 (\LF{cat}(x_1) \wedge \LF{run}(x_1))$ and $\exists x_7 (\LF{cat}(x_7) \wedge \LF{run}(x_7))$.

To address these shortcomings, successor benchmarks such as ReCOGS~\cite{wu-etal-2023-recogs} and SLOG~\cite{li-etal-2023-slog} adopt graph-matching-based metrics similar to Smatch, which account for permutations of conjuncts and variable renaming. ReCOGS in particular generates multiple logically equivalent variants of each target logical form, demonstrating that even minor surface differences (e.g., variable names or parentheses) can substantially affect evaluation. To our knowledge, however, none of these studies have employed a theorem prover to verify logical equivalence. A theorem prover naturally subsumes permutations of conjuncts and variable renaming as part of logical equivalence, and moreover is capable of handling richer equivalences involving, for example, negation and nested quantifiers.

Building on this background, the present study examines the capabilities of current neural semantic parsing models through evaluation with automated theorem proving. 
This line of inquiry is intended to lay the groundwork for developing models that achieve greater precision in natural language reasoning.

\section{Experimental Setup}
\label{section:experimental-setup}

This section describes the experimental setups for both SFT and ICL conducted in this study.

\subsection{Dataset}
\label{ssec:dataset}

We build our dataset from SICK~\citep{marelli2014sick}, a benchmark for natural language inference.
SICK consists of about 10,000 sentence pairs, each consisting of a premise (\texttt{p}) and a hypothesis (\texttt{h}), annotated with graded semantic relatedness (ranging from 1 to 5) as well as an entailment label chosen from $\setof{\mathtt{entailment}, \mathtt{contradiction}, \mathtt{neutral}}$. 
It includes linguistic phenomena such as quantification and negation, making it suitable for evaluating logically complex semantic representations.

We use its original train/test split for training and evaluation, strictly following the official division in all experiments. The training set contains 4,500 sentence pairs (9,000 sentences in total), and the test set contains 4,927 sentence pairs (9,854 sentences in total).

To obtain semantic representations that faithfully reflect sentence meaning, we parse all SICK sentences using ccg2lambda~\citep{mineshima-etal-2015-higher,martinez2016ccg2lambda}.
In this system, a CCG parser is first applied to produce a CCG derivation tree, which is then mapped into a logical form (semantic representation) via standard $\lambda$-calculus–based semantic composition; in our experiments, we use depccg~\citep{yoshikawa-etal-2017-ccg} as the CCG parser.
Given a premise $p$ and a hypothesis $h$, the system outputs their semantic representations $\mathrm{SR}(p)$ and $\mathrm{SR}(h)$, and then employs a theorem prover, together with axioms derived from external knowledge bases such as WordNet~\cite{fellbaum1998wordnet} to determine whether $\mathrm{SR}(p)$ entails $\mathrm{SR}(h)$, contradicts it, or neither.
The output is thus one of three labels: \texttt{entailment}, \texttt{contradiction}, or \texttt{neutral}.

For the target representation, we adopt event semantics~\citep{Parsons90}, a framework widely used in semantic parsing (e.g., in COGS~\cite{kim-linzen-2020-cogs}), and employ the event-semantics templates for ccg2lambda developed by \citet{martinez-gomez-etal-2017-demand}.
Table~\ref{tab:example-semantic} illustrates example sentences from SICK together with their event-semantic representation, where (b) shows the raw representations produced by ccg2lambda.

To obtain high-quality sentence-formula pairs, we filter SICK sentence pairs as follows.
We retain only those with a gold SICK label in $\setof{\mathtt{entailment}, \mathtt{contradiction}}$ for which the theorem prover's judgment over the ccg2lambda-derived representations matches the gold label.
Pairs labeled \texttt{neutral} are excluded, since such outcomes may occasionally arise from pipeline errors (e.g., parsing failures or misaligned logical forms), making it difficult to guarantee correct semantic representations.
From the retained pairs, we pair each sentence with its event-semantics formula to construct the training and evaluation instances. The resulting dataset comprises 2,392 training examples and 2,580 test examples.

We use two types of target representations to enable comparison with simplified formulas.
Table~\ref{tab:example-semantic} presents concrete examples of both representations, shown in (b) and (c), respectively.

\paragraph{1. Raw ccg2lambda outputs:} The unmodified output of ccg2lambda, represented as first-order predicate logic with event variables. In these formulas, quantifiers (particularly existential quantifiers) may appear in different positions depending on the sentence structure, and variable names are assigned arbitrarily.

\paragraph{2. Prenex-normalized formulas (PNF):} Derived deterministically from the raw \texttt{ccg2lambda} outputs by (i) moving all quantifiers to the sentence prefix  with systematic variable renaming (while keeping negations at the front of the sentence), (ii) normalizing predicate symbols (e.g., removing leading underscores introduced by \texttt{ccg2lambda}), and (iii) reassigning variable indices starting from 1. This normalization reduces incidental variation in quantifier placement and superficial symbol noise that could otherwise confound sequence models.

\medskip

We include both variants to assess the effect of target-side standardization via prenex normalization on model performance.

\begin{table*}[ht]
\centering
\caption{Example sentences annotated by category. 
\#Train indicates the number of occurrences in the SICK train data and \#Test indicates the number of occurrences in the SICK test data.
}
\begin{tabularx}{\textwidth}{l l l X}
\hline
\textbf{Category} & \textbf{\#Train} & \textbf{\#Test} &  \textbf{Example} \\
\hline
Conj & 1846 & 2065 & There is no dog wrestling and hugging. (\texttt{sick\_train\_13\_h}) \\
PP & 1500 & 1679 & A little girl is looking at a woman in costume. (\texttt{sick\_train\_74\_p}) \\
Passive & 795 & 839 & Children covered by leaves are playing with red shirts.  (\texttt{sick\_train\_61\_p}) \\
\hline
\end{tabularx}
\label{tab:category-examples}
\end{table*}

\subsection{Annotation}\label{subsec:annotation}
We introduced additional categories to assess which natural language phenomena present challenges for neural semantic parsing. Specifically, using the output of the CCG parser, we annotated the SICK dataset with three syntactic categories:

\begin{itemize}
    \item \textbf{Coordinating Conjunctions (CC):} Sentences containing coordinating conjunctions such as ``and'' and ``or'', which require proper handling of coordination scope.
    \item \textbf{Prepositional Phrases (PP):} Sentences with prepositional phrases that introduce spatial, temporal, or relational information requiring compositional analysis.
    \item \textbf{Passive Voice (PSS):} Sentences exhibiting passive voice constructions where argument structure differs from canonical active voice.
\end{itemize}

The annotation was conducted using automated extraction from CCG parse trees. We developed search scripts to identify these phenomena from syntactic categories and logical formulas patterns.

Table~\ref{tab:category-examples} shows the distribution of these categories in our dataset, with CC being the most frequent (2,790 instances), followed by PP (2,458 instances) and PSS (1,519 instances). This annotation enables us to analyze parser performance across different linguistic phenomena and identify which categories are particularly challenging for neural approaches.

\subsection{Supervised Fine-tuning}
To construct the semantic parser via SFT, we adopted T5-Small and T5-Base \citep{2020t5} as pretrained models. Training was conducted for 50 epochs with a batch size of 16, a learning rate of $1\times 10^{-5}$, weight decay of 0.01, and a warm-up of 500 steps. The maximum input and output sequence lengths were set to 256 tokens each. For each model–task combination, we trained and evaluated with seeds 1–10.

\begin{table}
\centering
\caption{
DRSs converted from the examples in Table~\ref{tab:example-semantic}:
\texttt{sick\_train\_88\_p} (left)
and
\texttt{sick\_train\_55\_p} (right).
}
\scalebox{0.9}{
\negdrs{e2 x1 x3}{
  biker(x1) \\
  jump(e2) \\
  subj(e2) = x1 \\
  air(x3) \\
  in(e2,x3)}
\hspace{2em}
\drs{e2 x1 x3}{
  boy(x1) \\
  three(x1) \\
  jump(e2) \\
  subj(e2) = x1 \\
  leaf(x3) \\
  in(e2,x3)}
} 
\label{tab:drs-conversion}
\end{table}

\subsection{In-context Learning}\label{subsection:icl}

We conducted few-shot in-context semantic parsing with large language models. For this experiment, we selected GPT-4o, GPT-4.1, and GPT-5, representing state-of-the-art models of different sizes that are widely used for reasoning-oriented tasks.
The prompt comprised five text–formula exemplars randomly sampled from the training set, together with basic conventions for the target formalism (e.g., notation for existential quantification and negation, and predicate-naming conventions for compound expressions). The full prompt is provided in Appendix~\ref{app:icl-prompt}.

\todo{For all models, the temperature was set to $0.0$. To account for variability, we ran each configuration with random seeds fixed at $1$, $2$, and $3$.} For GPT-5, we additionally set the API parameters \texttt{reasoning.effort} (a budget indicator for the reasoning phase) and \texttt{text.verbosity} (the verbosity of the textual response) to \texttt{minimal}.

For cost considerations related to API usage, we restricted ICL to prenex-normalized data and did not include raw ccg2lambda outputs, and we evaluated each model in a single run.

\begin{table*}[t]
  \centering
  \small
  \setlength{\tabcolsep}{5pt}
  \caption{SFT results (raw ccg2lambda outputs): mean $\pm$ standard deviation ($n{=}10$)}
  \label{tab:ft-raw}
  \begin{tabular}{lcccccc}
    \toprule
    \textbf{Model} & \textbf{Exact Match} & \textbf{Prover Acc} & \textbf{Dmatch Precision} & \textbf{Dmatch Recall} & \textbf{Dmatch F1} & \textbf{Non-WFF Ratio} \\
    \midrule
    T5-Small & \(0.101 \pm 0.003\) & \(0.189 \pm 0.005\) & \(0.611 \pm 0.011\) & \(0.504 \pm 0.010\) & \(0.544 \pm 0.010\) & \(0.240 \pm 0.012\) \\
    T5-Base & \(0.322 \pm 0.002\) & \(0.634 \pm 0.004\) & \(0.887 \pm 0.004\) & \(0.864 \pm 0.004\) & \(0.873 \pm 0.004\) & \(0.031 \pm 0.003\) \\
    \bottomrule
  \end{tabular}
\end{table*}

\begin{table*}[t]
  \centering
  \small
  \setlength{\tabcolsep}{5pt}
  \caption{SFT results (prenex-normalized formulas): mean $\pm$ standard deviation ($n{=}10$)}
  \label{tab:ft-pnf}
  \begin{tabular}{lcccccc}
    \toprule
    \textbf{Model} & \textbf{Exact Match} & \textbf{Prover Acc} & \textbf{Dmatch Precision} & \textbf{Dmatch Recall} & \textbf{Dmatch F1} & \textbf{Non-WFF Ratio} \\
    \midrule
    T5-Small & \(0.411 \pm 0.003\) & \(0.439 \pm 0.002\) & \(0.771 \pm 0.002\) & \(0.739 \pm 0.002\) & \(0.752 \pm 0.002\) & \(0.018 \pm 0.002\) \\
    T5-Base & \(0.674 \pm 0.004\) & \(0.689 \pm 0.004\) & \(0.889 \pm 0.002\) & \(0.874 \pm 0.002\) & \(0.880 \pm 0.002\) & \(0.007 \pm 0.001\) \\
    \bottomrule
  \end{tabular}
\end{table*}

\subsection{Evaluation Metrics}
We compare the semantic parsing outputs of each method using two evaluation metrics: graph-matching and automated theorem proving.

For the graph-matching evaluation, we use Counter \cite{van-noord-etal-2018-evaluating}.\footnote{\url{https://github.com/RikVN/DRS_parsing}} Counter is a modification of Smatch for graph structures in which scope-taking phenomena such as negation and quantification matter. 
\todo{Because Counter supports DRSs, we convert the parser's predictions (i.e., FOL formulas in event semantics) into DRSs and use Counter to compute the F-score between the gold DRS and the predicted DRS.
We refer to this F-score, together with its components precision and recall, as \textit{Dmatch}.
The conversion from FOL to DRS was performed using the conversion script provided in ccg2lambda.\footnote{\url{https://github.com/mynlp/ccg2lambda}}
For example, the two formulas in Table~\ref{tab:example-semantic} are converted into DRSs, as shown in Table~\ref{tab:drs-conversion}.
}\footnote{The indices of the variables for entities and events are assigned according to the order of their occurrences in the Raw formula in Table~\ref{tab:example-semantic}.
}

In addition, we evaluate via automated theorem proving. We use Vampire \cite{kovacs2013first}, a state-of-the-art first-order theorem prover.\footnote{\url{https://github.com/vprover/vampire}} To examine whether the parser's prediction is logically equivalent to the gold reference, we test whether bidirectional entailment holds between the gold formula and the prediction.

Dmatch provides a similarity score based on clause overlap in DRSs, but it does not reveal the precise logical relation between two formulas (e.g., equivalence, entailment, or contradiction).
For example, consider the following pairs:

\medskip

$g_1 =$ \texttt{exists\allowbreak{} e\allowbreak{}.jump(e)}

$p_1 =$ \texttt{exists\allowbreak{} e\allowbreak{}.(jump(e) \& high(e))}

\medskip

$g_2 =$ \texttt{exists\allowbreak{} e\allowbreak{} x.(eat(e) \& (subj(e)=x))}

$p_2 =$ \texttt{exists\allowbreak{} e\allowbreak{} x.(eat(e) \& (obj(e)=x))}

\medskip

\noindent
Both $\tuple{g_1, p_1}$ and $\tuple{g_2, p_2}$
receive the same Dmatch score of 0.5, even though the first reflects one-way entailment (the prediction $p_1$ over-specifies the gold $g_1$) while the second is a clear semantic role mismatch with no entailment.
Similarly, the pair of logically unrelated formulas $\LF{P(a)}$ and $\LF{Q(a)}$, and the contradictory pair $\LF{P(a)}$ and $\LF{-P(a)}$, both receive a Dmatch score of 0 and are thus indistinguishable.
Since such distinctions are essential for reasoning tasks, we complement graph-based evaluation with theorem proving, which explicitly identifies logical relations.

\begin{figure*}[ht]
  \centering
  \includegraphics[width=\linewidth]{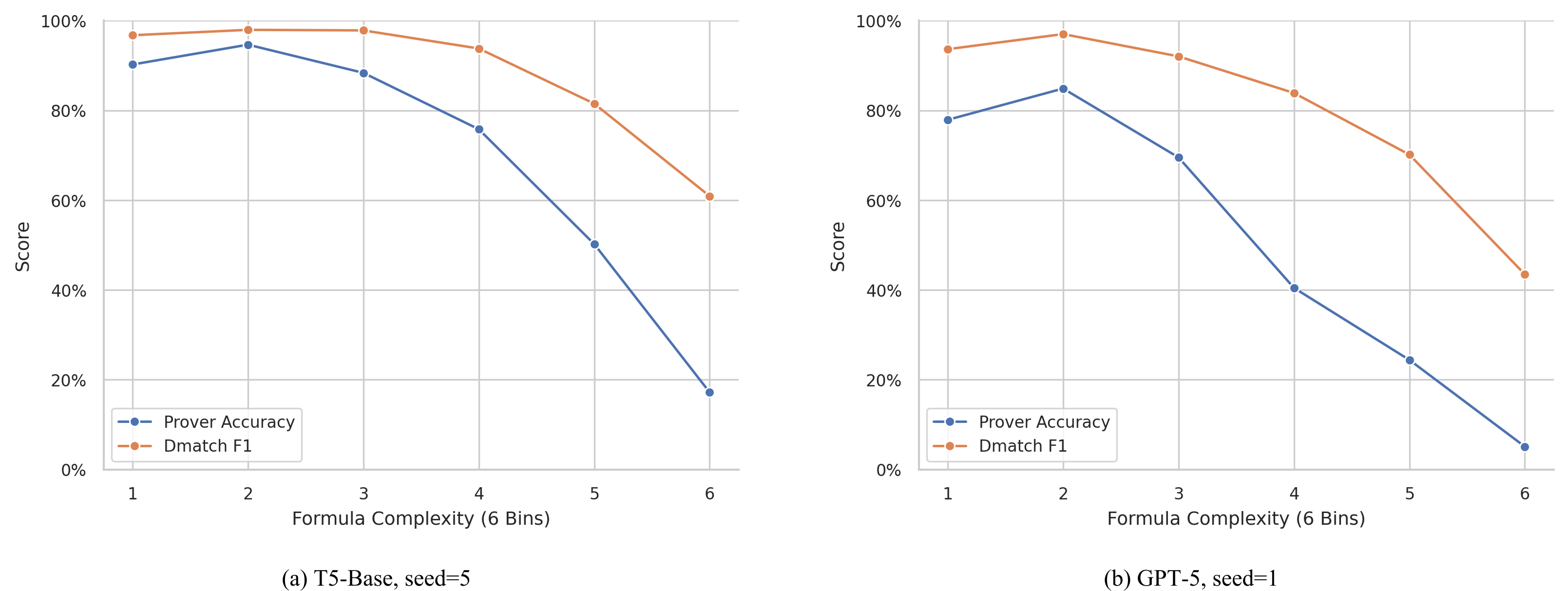}
  \caption{Relationship between target-side formula complexity and model performance. (a) best SFT configuration (T5-Base, seed 5); (b) best ICL configuration (GPT-5, seed 1).}
  \label{fig:complexity}
\end{figure*}

\section{Results}\label{section:results}

We report results for SFT and few-shot ICL on our semantic parsing task, evaluated using three criteria: graph-matching (Dmatch), theorem-prover–based equivalence, and well-formedness. All metrics are reported to three decimal places (rounded half-up).

\subsection{Supervised Fine-tuning}
For SFT, we trained with random seeds 1–10 on two target representations (raw formulas and prenex formulas) and evaluated on the test split. Tables~\ref{tab:ft-raw} and \ref{tab:ft-pnf} report the results.

Across both representations, T5-Base shows higher values than T5-Small. The gaps in Prover Accuracy and Dmatch F1 are present in both settings and are larger on raw ccg2lambda outputs; the Non-WFF Ratio is lower for T5-Base in both settings, especially on raw. 

Within each model, prenex normalization is associated with higher Prover Accuracy and Exact Match and a lower Non-WFF Ratio. This pattern indicates that suppressing incidental variability in the target-side quantificational structure is associated with better parser performance. 
 
The results therefore point to concrete directions for improving semantic parsers: first, leverage natural scaling with model size---performance increases consistently from T5-Small to T5-Base across both representations and metrics---indicating that greater capacity is a straightforward path to better end-to-end behavior; second, normalize the target representation via PNF to suppress incidental variability, reduce the Non-WFF ratio, and improve alignment with theorem-proving-based evaluation.

Looking across both representations and both models, higher Dmatch F1 does not translate into commensurately high Prover Accuracy. For raw ccg2lambda outputs, T5-Small attains Dmatch F1 of 0.544, whereas Prover Accuracy is 0.189; even T5-Base shows 0.873 and 0.634, respectively. With prenex normalization the gap narrows but remains (0.752 and 0.439 for T5-Small; 0.880 and 0.689 for T5-Base). This consistent disparity indicates that clause-level graph-matching captures structural overlap rather than the logical equivalence, underscoring the need for evaluation that are sensitive to logical structure.

\begin{table*}[t]
  \centering
  \small
  \setlength{\tabcolsep}{5pt}
  \caption{ICL results (prenex-normalized formulas): mean $\pm$ standard deviation ($n{=}3$)}
  \label{tab:icl-raw}
  \begin{tabular}{lcccccc}
    \toprule
    \textbf{Model} & \textbf{Exact Match} & \textbf{Prover Acc} & \textbf{Dmatch Precision} & \textbf{Dmatch Recall} & \textbf{Dmatch F1} & \textbf{Non-WFF Ratio} \\
    \midrule
    GPT-4o  & \(0.328\pm0.029\) & \(0.493\pm0.046\) & \(0.824\pm0.017\) & \(0.812\pm0.026\) & \(0.816\pm0.022\) & \(0.012\pm0.006\) \\
    GPT-4.1 & \(0.278\pm0.046\) & \(0.474\pm0.034\) & \(0.734\pm0.027\) & \(0.737\pm0.033\) & \(0.733\pm0.030\) & \(0.083\pm0.025\) \\
    GPT-5   & \(0.318\pm0.037\) & \(0.514\pm0.009\) & \(0.806\pm0.013\) & \(0.803\pm0.010\) & \(0.803\pm0.011\) & \(0.010\pm0.002\) \\
    \bottomrule
  \end{tabular}
\end{table*}

\begin{figure*}[t]
  \centering
  \includegraphics[width=\linewidth]{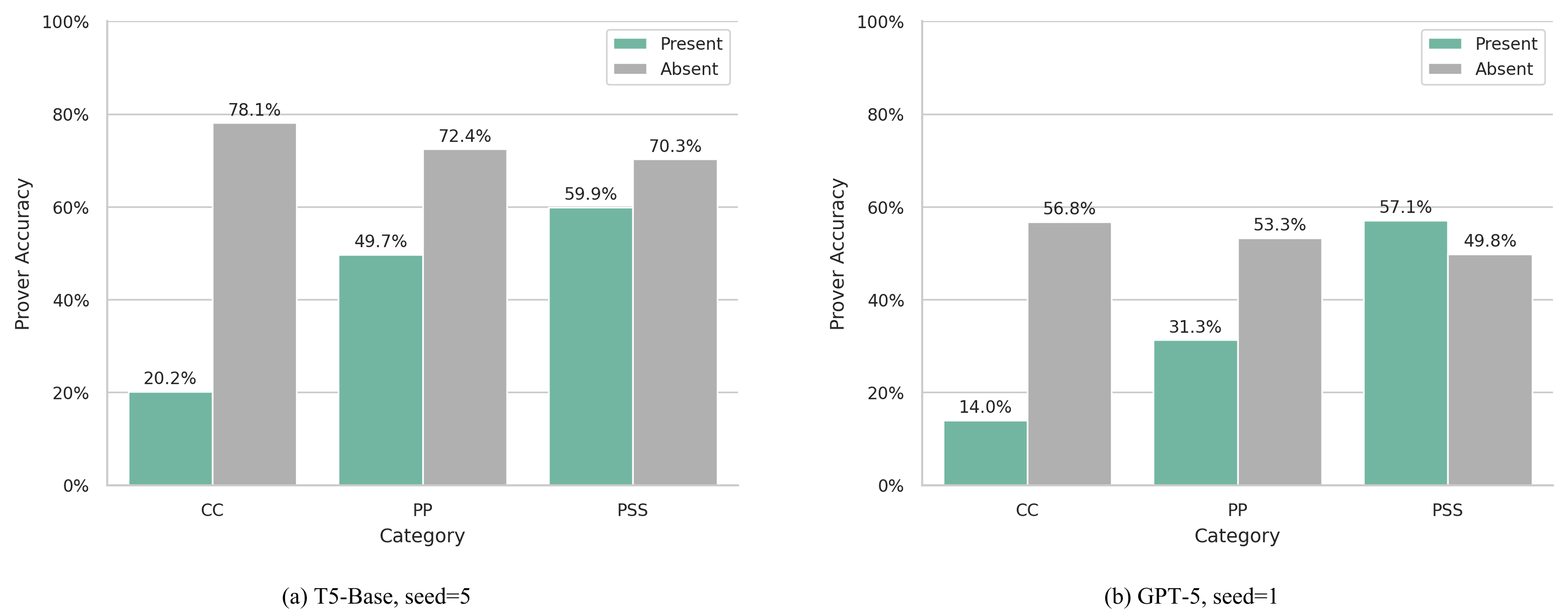}
  \caption{Prover accuracy stratified by the presence or absence of syntactic features. X-axis labels follow Section~\ref{subsec:annotation}. (a) best SFT configuration (T5-Base, seed 5); (b) best ICL configuration (GPT-5, seed 1).}
  \label{tab:struct-results}
\end{figure*}

\subsection{In-context Learning}
\todo{All ICL results are obtained on prenex-normalized targets and averaged over three runs with random seeds $1$, $2$, and $3$ (Table~\ref{tab:icl-raw}). GPT-5 achieves the best overall scores (Exact Match 0.318; Prover Accuracy 0.514; Dmatch F1 0.803). GPT-4o follows (0.328; 0.493; 0.816), and GPT-4.1 shows lower Dmatch F1 (0.733) and the highest Non-WFF Ratio (0.083). Both GPT-4o and GPT-5 maintain a very low Non-WFF Ratio (0.012 and 0.010), indicating that the prompt conventions yield mostly well-formed formulas.}

\todo{Relative to SFT on prenex targets (Table~\ref{tab:ft-pnf}), even the strongest ICL setting (GPT-5) trails T5-Base in Prover Accuracy (0.514 compared with 0.689) and Exact Match (0.318 compared with 0.674), while matching well-formedness (Non-WFF Ratio 0.010 compared with 0.007). Dmatch F1 is lower than T5-Base (0.803 compared with 0.880) but remains in a similar range; as noted in the SFT results, graph-matching scores tend to run higher than prover-based equivalence.}

\section{Discussion and Future Perspectives}
\todo{In this section, we focus our analysis on the strongest configuration within SFT---T5-Base (seed 5, final checkpoint)---and, for ICL, we use GPT-5 with seed 1 as a representative setting. All results are restricted to prenex-normalized targets.} We begin by examining how target-side formula complexity modulates performance, then analyze the impact of syntactic and semantic phenomena (coordinating conjunctions, prepositional phrases, passive voice), and finally investigate the sources of mispredictions through a category-wise breakdown centered on the overall best model, T5-Base.

\subsection{Impact of Formula Complexity}
We grouped test instances by the complexity of the target-side formula and computed Prover Accuracy and Dmatch for each bin. 
Formula complexity was measured by counting logical constants, specifically negation, quantifiers, and conjunctions (see Table~\ref{tab:example-semantic} for examples). 
Instances were sorted by complexity and split into six equal-sized groups. In our test set ($n=2,580$), each group therefore contains 430 instances.
Results in Figure~\ref{fig:complexity} show a clear performance drop as complexity increases, with T5-Base consistently outperforming GPT-5 across all bins. 
Notably, even GPT-5, despite its ability to handle very long outputs (up to 128k tokens), remains sensitive to structural complexity, suggesting that future work should focus on improving robustness to long and compositional formulas.

\subsection{Impact of Syntactic Features}
We conducted a stratified analysis by linguistic phenomena, comparing Prover Accuracy between the presence and absence of coordination, prepositional phrases, and passive voice. 
Figure~\ref{tab:struct-results} shows the results.
Table~\ref{tab:nsp-errors} in Appendix~\ref{appendix:error-examples} shows error examples of each category.
The results show that coordination is a major source of difficulty in both settings: its presence is associated with a large reduction in Prover Accuracy relative to its absence. Prepositional phrases also correlate with decreased accuracy, reflecting the burden of resolving relational structure and attachment in a way that remains faithful to subsequent formal inference. By contrast, passive voice exhibits divergent behavior across models: one model appears less sensitive or even slightly advantaged by passive constructions, whereas the other shows decreased accuracy under passive voice. A plausible interpretation is that the prompt and large pretraining may induce templates that better regularize argument-structure alternations, while fine-tuned parameters benefit more from canonical (active) realizations emphasized during training.

\begin{figure}[t]
  \centering
  \includegraphics[width=\linewidth]{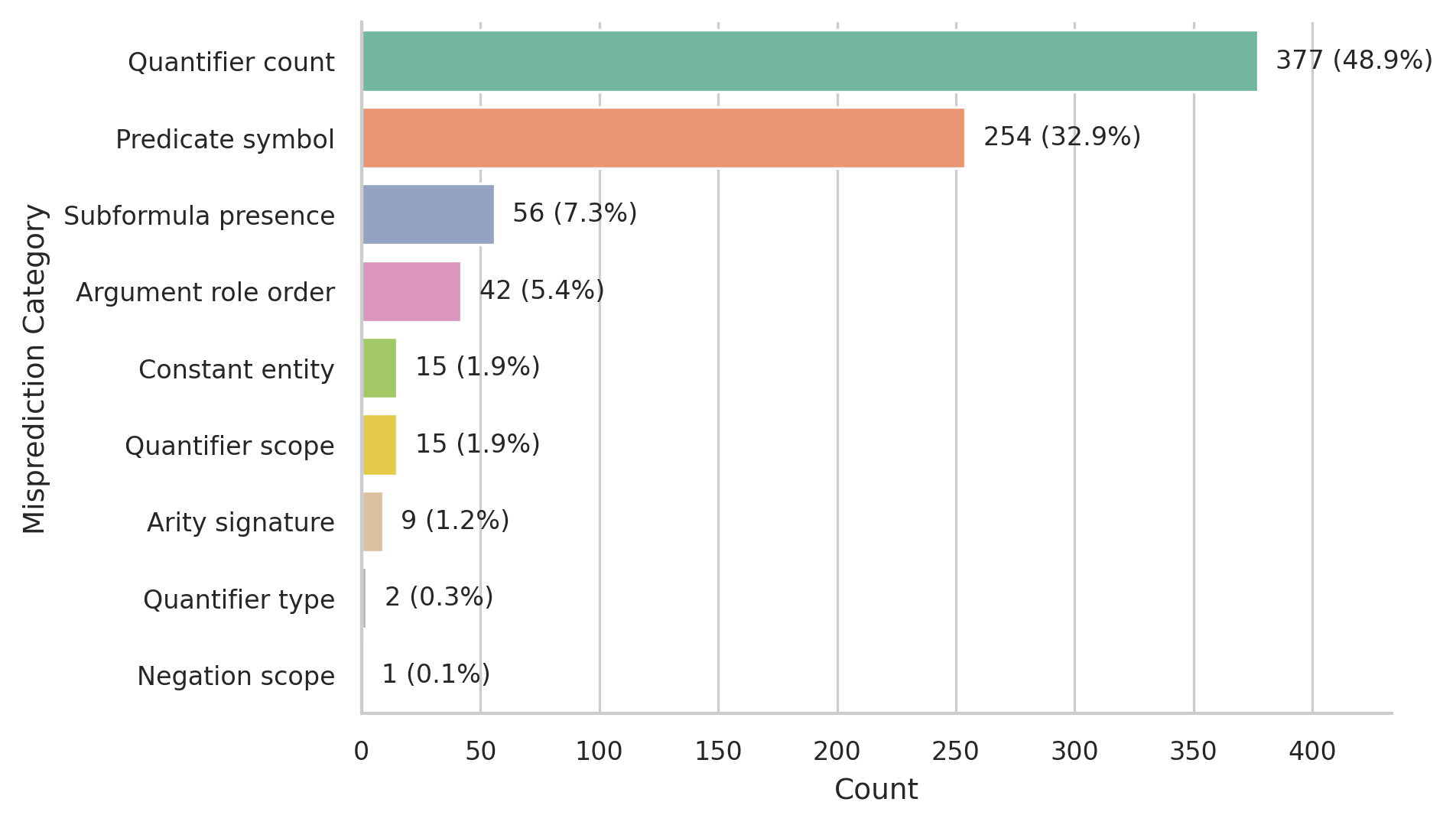}
  \caption{Error type distribution of 771 prover-failed predictions by T5-Base.}
  \label{fig:midpred-classify}
\end{figure}

\subsection{Breakdown of Mispredictions}
Among the predictions of T5-Base, the best overall model, we selected 771 cases that were well-formed but failed the prover-based evaluation. 
For fine-grained error analysis, these cases were presented individually to an LLM (GPT-4o), which was instructed to assign a label corresponding to the most critical problem in each prediction. 
The set of labels was predefined by the authors, and the LLM selected one from this set. 
In total, we prepared 11 label types, including, for example, Predicate Symbols (predicate mismatches such as \LF{loud} vs.\ \LF{loudly}), Subformula Presence (missing subformulas), and Argument Role Order (incorrect variable order in \LF{subj} and \LF{obj} functions).

The results are summarized in Figure~\ref{fig:midpred-classify}. The two most frequent error types, quantifier-count mismatch and predicate-name error, together account for 81.8\% of all mispredictions.
The most frequent category, quantifier-count mismatch, accounts for 48.9\% overall. 
Table~\ref{tab:quantifier-errors} in Appendix ~\ref{app:quantifier-error-examples} shows two representative examples.
These errors are not superficial notation issues; they reflect fundamental misinterpretations of the semantic structure of the source text. Addressing them requires strategies that tie natural language understanding more tightly to the intended semantic framework (here, event semantics) during prediction.

\subsection{Outlook Based on the Behaviors of SFT and ICL}
Based on the respective behaviors of SFT and ICL, a coherent picture emerges: parameter learning aligns outputs more tightly with theorem-prover criteria, whereas prompt-only conditioning of foundation models yields consistently well-formed formulas under prenex conventions yet comparatively weaker semantic alignment. This suggests a practical division of labor---use ICL as a high–well-formedness candidate generator, paired with a component optimized for semantic adequacy (e.g., a fine-tuned parser or a prover-guided reranker/repair module). Such a hybrid pipeline can narrow the remaining gap without sacrificing the strengths of either approach.

\section{Conclusion}
We propose evaluation with an automated theorem prover as a precise test of whether a neural semantic parser faithfully captures a sentence's semantic structure. Coupled with graph-matching, it reveals a persistent gap between surface overlap and logical equivalence across supervised fine-tuning and in-context learning, with normalization improving well formedness and logical adequacy. Sentences with coordination, prepositional phrases, or passive voice are more error prone, and errors concentrate in variable binding and indexing and in predicate naming. Looking ahead, a hybrid strategy that pairs generation with logic-aware verification is a promising direction.

\section{Limitations}

Although we employ popular approaches such as SFT with T5 and ICL with OpenAI models, we have not, for example, fine-tuned billion-parameter variants or evaluated architectures with substantially different design principles; accordingly, the applicability of our conclusions is limited. Additionally, we use OpenAI models via API access, which limits transparency into the models and may constrain long-term reproducibility.

Our analysis is conducted on SICK, a natural language inference benchmark that is widely used for studying entailment relations involving compositional logical structures such as negation and quantification; we chose this dataset for precisely these properties, while noting that examining additional datasets and domains will help assess generality.

Another limitation of this study is that we do not evaluate end-to-end natural language inference, one of the principal downstream tasks. Instead, we focus on bidirectional entailment between predicted and gold logical forms, since this directly serves our objective of assessing the formulas themselves.

\section*{Acknowledgements}

We would like to thank the anonymous reviewers for their helpful comments and suggestions.
This work is partially supported by JST, CREST Grant Number JPMJCR2114.

\bibliography{j_yourrefs}

\begin{thebibliography}{39}
\providecommand{\natexlab}[1]{#1}

\bibitem[{Abzianidze(2015)}]{abzianidze:2015:EMNLP}
Lasha Abzianidze. 2015.
\newblock \href {http://aclweb.org/anthology/D15-1296} {A tableau prover for
  natural logic and language}.
\newblock In \emph{Proceedings of the 2015 Conference on Empirical Methods in
  Natural Language Processing}, pages 2492--2502, Lisbon, Portugal.

\bibitem[{Abzianidze et~al.(2017)Abzianidze, Bjerva, Evang, Haagsma, van Noord,
  Ludmann, Nguyen, and Bos}]{abzianidze-etal-2017-parallel}
Lasha Abzianidze, Johannes Bjerva, Kilian Evang, Hessel Haagsma, Rik van Noord,
  Pierre Ludmann, Duc-Duy Nguyen, and Johan Bos. 2017.
\newblock \href {https://aclanthology.org/E17-2039/} {The {P}arallel {M}eaning
  {B}ank: Towards a multilingual corpus of translations annotated with
  compositional meaning representations}.
\newblock In \emph{Proceedings of the 15th Conference of the {E}uropean Chapter
  of the Association for Computational Linguistics: Volume 2, Short Papers},
  pages 242--247.

\bibitem[{Bai et~al.(2022)Bai, Chen, and Zhang}]{bai2022graph}
Xuefeng Bai, Yulong Chen, and Yue Zhang. 2022.
\newblock Graph pre-training for {AMR} parsing and generation.
\newblock \emph{arXiv preprint arXiv:2203.07836}.

\bibitem[{Banarescu et~al.(2013)Banarescu, Bonial, Cai, Georgescu, Griffitt,
  Hermjakob, Knight, Koehn, Palmer, and
  Schneider}]{banarescu-etal-2013-abstract}
Laura Banarescu, Claire Bonial, Shu Cai, Madalina Georgescu, Kira Griffitt, Ulf
  Hermjakob, Kevin Knight, Philipp Koehn, Martha Palmer, and Nathan Schneider.
  2013.
\newblock \href {https://aclanthology.org/W13-2322/} {{A}bstract {M}eaning
  {R}epresentation for sembanking}.
\newblock In \emph{Proceedings of the 7th Linguistic Annotation Workshop and
  Interoperability with Discourse}, pages 178--186.

\bibitem[{Blackburn and Bos(2005)}]{blackburn2005representation}
Patrick Blackburn and Johan Bos. 2005.
\newblock \emph{Representation and {I}nference for {N}atural {L}anguage: {A}
  {F}irst {C}ourse in {C}omputational {S}emantics}.
\newblock CSLI.

\bibitem[{Bos et~al.(2004)Bos, Clark, Steedman, Curran, and
  Hockenmaier}]{bos2004wide}
Johan Bos, Stephen Clark, Mark Steedman, James~R Curran, and Julia Hockenmaier.
  2004.
\newblock Wide-coverage semantic representations from a {CCG} parser.
\newblock In \emph{Proceedings of the 20th international conference on
  Computational Linguistics}, pages 1240--1246.

\bibitem[{Bos and Markert(2005)}]{bos-markert-2005-recognising}
Johan Bos and Katja Markert. 2005.
\newblock \href {https://aclanthology.org/H05-1079/} {Recognising textual
  entailment with logical inference}.
\newblock In \emph{Proceedings of Human Language Technology Conference and
  Conference on Empirical Methods in Natural Language Processing}, pages
  628--635.

\bibitem[{Brown et~al.(2020)Brown, Mann, Ryder, Subbiah, Kaplan, Dhariwal,
  Neelakantan, Shyam, Sastry, Askell, Agarwal, Herbert-Voss, Krueger, Henighan,
  Child, Ramesh, Ziegler, Wu, Winter, Hesse, Chen, Sigler, Litwin, Gray, Chess,
  Clark, Berner, McCandlish, Radford, Sutskever, and
  Amodei}]{brown2020language}
Tom Brown, Benjamin Mann, Nick Ryder, Melanie Subbiah, Jared Kaplan, Prafulla
  Dhariwal, Arvind Neelakantan, Pranav Shyam, Girish Sastry, Amanda Askell,
  Sandhini Agarwal, Ariel Herbert-Voss, Gretchen Krueger, Tom Henighan, Rewon
  Child, Aditya Ramesh, Daniel Ziegler, Jeffrey Wu, Clemens Winter, and 12
  others. 2020.
\newblock \href {https://arxiv.org/abs/2005.14165} {Language models are
  few-shot learners}.
\newblock In \emph{Advances in Neural Information Processing Systems},
  volume~33, pages 1877--1901.

\bibitem[{Cai and Knight(2013)}]{cai2013smatch}
Shu Cai and Kevin Knight. 2013.
\newblock Smatch: an evaluation metric for semantic feature structures.
\newblock In \emph{Proceedings of the 51st Annual Meeting of the Association
  for Computational Linguistics (Volume 2: Short Papers)}, pages 748--752.

\bibitem[{Devlin et~al.(2019)Devlin, Chang, Lee, and
  Toutanova}]{devlin2019bert}
Jacob Devlin, Ming-Wei Chang, Kenton Lee, and Kristina Toutanova. 2019.
\newblock \href {https://doi.org/10.18653/v1/N19-1423} {{BERT}: Pre-training of
  deep bidirectional transformers for language understanding}.
\newblock In \emph{Proceedings of the 2019 Conference of the North {A}merican
  Chapter of the Association for Computational Linguistics: Human Language
  Technologies, Volume 1 (Long and Short Papers)}, pages 4171--4186,
  Minneapolis, Minnesota. Association for Computational Linguistics.

\bibitem[{Dong and Lapata(2016)}]{dong-lapata-2016-language}
Li~Dong and Mirella Lapata. 2016.
\newblock \href {https://doi.org/10.18653/v1/P16-1004} {Language to logical
  form with neural attention}.
\newblock In \emph{Proceedings of the 54th Annual Meeting of the Association
  for Computational Linguistics (Volume 1: Long Papers)}, pages 33--43.

\bibitem[{Dong et~al.(2015)Dong, Wei, Liu, Zhou, and Xu}]{dong2015statistical}
Li~Dong, Furu Wei, Shujie Liu, Ming Zhou, and Ke~Xu. 2015.
\newblock A statistical parsing framework for sentiment classification.
\newblock \emph{Computational Linguistics}, 41(2):293--336.

\bibitem[{Fellbaum(1998)}]{fellbaum1998wordnet}
Christiane Fellbaum. 1998.
\newblock \emph{Word{N}et: An electronic lexical database}.
\newblock MIT press.

\bibitem[{Fitting(1996)}]{fitting1996first}
Melvin Fitting. 1996.
\newblock \emph{First-Order Logic and Automated Theorem Proving}.
\newblock Springer.

\bibitem[{Haruta et~al.(2022)Haruta, Mineshima, and
  Bekki}]{Haruta_Mineshima_Bekki_2022}
Izumi Haruta, Koji Mineshima, and Daisuke Bekki. 2022.
\newblock \href {https://doi.org/10.15398/jlm.v10i1.294} {Implementing natural
  language inference for comparatives}.
\newblock \emph{Journal of Language Modelling}, 10(1):139–191.

\bibitem[{Hockenmaier and Steedman(2007)}]{hockenmaier2007ccgbank}
Julia Hockenmaier and Mark Steedman. 2007.
\newblock {CCG}bank: a corpus of {CCG} derivations and dependency structures
  extracted from the {P}enn {T}reebank.
\newblock \emph{Computational Linguistics}, 33(3):355--396.

\bibitem[{Kautz(2022)}]{kautz2022third}
Henry~A Kautz. 2022.
\newblock The third {AI} summer: {AAAI} {R}obert {S}. {E}ngelmore {M}emorial
  {L}ecture.
\newblock \emph{AI Magazine}, 43(1).

\bibitem[{Kim and Linzen(2020)}]{kim-linzen-2020-cogs}
Najoung Kim and Tal Linzen. 2020.
\newblock \href {https://doi.org/10.18653/v1/2020.emnlp-main.731} {{COGS}: A
  compositional generalization challenge based on semantic interpretation}.
\newblock In \emph{Proceedings of the 2020 Conference on Empirical Methods in
  Natural Language Processing (EMNLP)}, pages 9087--9105, Online. Association
  for Computational Linguistics.

\bibitem[{Konstas et~al.(2017)Konstas, Iyer, Yatskar, Choi, and
  Zettlemoyer}]{konstas-etal-2017-neural}
Ioannis Konstas, Srinivasan Iyer, Mark Yatskar, Yejin Choi, and Luke
  Zettlemoyer. 2017.
\newblock \href {https://doi.org/10.18653/v1/P17-1014} {Neural {AMR}:
  Sequence-to-sequence models for parsing and generation}.
\newblock In \emph{Proceedings of the 55th Annual Meeting of the Association
  for Computational Linguistics (Volume 1: Long Papers)}, pages 146--157.

\bibitem[{Kov{\'a}cs and Voronkov(2013)}]{kovacs2013first}
Laura Kov{\'a}cs and Andrei Voronkov. 2013.
\newblock First-order theorem proving and vampire.
\newblock In \emph{International Conference on Computer Aided Verification},
  pages 1--35. Springer.

\bibitem[{Lepore and Cumming(2009)}]{lepore2009}
Ernest Lepore and Sam Cumming. 2009.
\newblock \emph{Meaning and Argument: An Introduction to Logic Through
  Language}.
\newblock Wiley-Blackwell.

\bibitem[{Li et~al.(2023)Li, Donatelli, Koller, Linzen, Yao, and
  Kim}]{li-etal-2023-slog}
Bingzhi Li, Lucia Donatelli, Alexander Koller, Tal Linzen, Yuekun Yao, and
  Najoung Kim. 2023.
\newblock \href {https://doi.org/10.18653/v1/2023.emnlp-main.194} {{SLOG}: A
  structural generalization benchmark for semantic parsing}.
\newblock In \emph{Proceedings of the 2023 Conference on Empirical Methods in
  Natural Language Processing}, pages 3213--3232, Singapore. Association for
  Computational Linguistics.

\bibitem[{Ling et~al.(2016)Ling, Blunsom, Grefenstette, Hermann,
  Ko{\v{c}}isk{\'y}, Wang, and Senior}]{ling-etal-2016-latent}
Wang Ling, Phil Blunsom, Edward Grefenstette, Karl~Moritz Hermann,
  Tom{\'a}{\v{s}} Ko{\v{c}}isk{\'y}, Fumin Wang, and Andrew Senior. 2016.
\newblock \href {https://doi.org/10.18653/v1/P16-1057} {Latent predictor
  networks for code generation}.
\newblock In \emph{Proceedings of the 54th Annual Meeting of the Association
  for Computational Linguistics (Volume 1: Long Papers)}, pages 599--609.

\bibitem[{Marelli et~al.(2014)Marelli, Menini, Baroni, Bentivogli, Bernardi,
  and Zamparelli}]{marelli2014sick}
Marco Marelli, Stefano Menini, Marco Baroni, Luisa Bentivogli, Raffaella
  Bernardi, and Roberto Zamparelli. 2014.
\newblock A sick cure for the evaluation of compositional distributional
  semantic models.
\newblock In \emph{Proceedings of the Ninth International Conference on
  Language Resources and Evaluation (LREC'14)}, pages 216--223.

\bibitem[{Mart{\'\i}nez-G{\'o}mez et~al.(2016)Mart{\'\i}nez-G{\'o}mez,
  Mineshima, Miyao, and Bekki}]{martinez2016ccg2lambda}
Pascual Mart{\'\i}nez-G{\'o}mez, Koji Mineshima, Yusuke Miyao, and Daisuke
  Bekki. 2016.
\newblock ccg2lambda: A compositional semantics system.
\newblock In \emph{Proceedings of ACL-2016 System Demonstrations}, pages
  85--90.

\bibitem[{Mart{\'i}nez-G{\'o}mez et~al.(2017)Mart{\'i}nez-G{\'o}mez, Mineshima,
  Miyao, and Bekki}]{martinez-gomez-etal-2017-demand}
Pascual Mart{\'i}nez-G{\'o}mez, Koji Mineshima, Yusuke Miyao, and Daisuke
  Bekki. 2017.
\newblock \href {https://aclanthology.org/E17-1067/} {On-demand injection of
  lexical knowledge for recognising textual entailment}.
\newblock In \emph{Proceedings of the 15th Conference of the {E}uropean Chapter
  of the Association for Computational Linguistics: Volume 1, Long Papers},
  pages 710--720, Valencia, Spain. Association for Computational Linguistics.

\bibitem[{Mineshima et~al.(2015)Mineshima, Mart{\'i}nez-G{\'o}mez, Miyao, and
  Bekki}]{mineshima-etal-2015-higher}
Koji Mineshima, Pascual Mart{\'i}nez-G{\'o}mez, Yusuke Miyao, and Daisuke
  Bekki. 2015.
\newblock \href {https://doi.org/10.18653/v1/D15-1244} {Higher-order logical
  inference with compositional semantics}.
\newblock In \emph{Proceedings of the 2015 Conference on Empirical Methods in
  Natural Language Processing}, pages 2055--2061.

\bibitem[{Opitz(2023)}]{opitz-2023-smatch}
Juri Opitz. 2023.
\newblock \href {https://doi.org/10.18653/v1/2023.findings-eacl.118}
  {{SMATCH}++: Standardized and extended evaluation of semantic graphs}.
\newblock In \emph{Findings of the Association for Computational Linguistics:
  EACL 2023}, pages 1595--1607, Dubrovnik, Croatia. Association for
  Computational Linguistics.

\bibitem[{Opitz et~al.(2020)Opitz, Parcalabescu, and
  Frank}]{opitz-etal-2020-amr}
Juri Opitz, Letitia Parcalabescu, and Anette Frank. 2020.
\newblock \href {https://doi.org/10.1162/tacl_a_00329} {{AMR} similarity
  metrics from principles}.
\newblock \emph{Transactions of the Association for Computational Linguistics},
  8:522--538.

\bibitem[{Parsons(1990)}]{Parsons90}
Terence Parsons. 1990.
\newblock \emph{Events in the Semantics of {E}nglish}.
\newblock MIT Press, Cambridge, MA.

\bibitem[{Raffel et~al.(2020)Raffel, Shazeer, Roberts, Lee, Narang, Matena,
  Zhou, Li, and Liu}]{2020t5}
Colin Raffel, Noam Shazeer, Adam Roberts, Katherine Lee, Sharan Narang, Michael
  Matena, Yanqi Zhou, Wei Li, and Peter~J. Liu. 2020.
\newblock \href {http://jmlr.org/papers/v21/20-074.html} {Exploring the limits
  of transfer learning with a unified text-to-text transformer}.
\newblock \emph{Journal of Machine Learning Research}, 21(140):1--67.

\bibitem[{Robinson and Voronkov(2001)}]{robinson2001handbook}
Alan Robinson and Andrei Voronkov. 2001.
\newblock \emph{Handbook of Automated Reasoning}, volume~1.
\newblock Elsevier.

\bibitem[{Steedman(2000)}]{Steedman2000}
Mark~J. Steedman. 2000.
\newblock \emph{The Syntactic Process}.
\newblock MIT Press, Cambridge.

\bibitem[{van Noord et~al.(2018)van Noord, Abzianidze, Haagsma, and
  Bos}]{van-noord-etal-2018-evaluating}
Rik van Noord, Lasha Abzianidze, Hessel Haagsma, and Johan Bos. 2018.
\newblock \href {https://aclanthology.org/L18-1267/} {Evaluating scoped meaning
  representations}.
\newblock In \emph{Proceedings of the Eleventh International Conference on
  Language Resources and Evaluation ({LREC} 2018)}. European Language Resources
  Association (ELRA).

\bibitem[{van Noord et~al.(2020)van Noord, Toral, and
  Bos}]{van-noord-etal-2020-character}
Rik van Noord, Antonio Toral, and Johan Bos. 2020.
\newblock \href {https://doi.org/10.18653/v1/2020.emnlp-main.371}
  {Character-level representations improve {DRS}-based semantic parsing even in
  the age of {BERT}}.
\newblock In \emph{Proceedings of the 2020 Conference on Empirical Methods in
  Natural Language Processing (EMNLP)}, pages 4587--4603.

\bibitem[{Wu et~al.(2022)Wu, Jiang, Li, Rabe, Staats, Jamnik, and
  Szegedy}]{wu2022autoformalization}
Yuhuai Wu, Albert~Qiaochu Jiang, Wenda Li, Markus Rabe, Charles Staats, Mateja
  Jamnik, and Christian Szegedy. 2022.
\newblock Autoformalization with large language models.
\newblock \emph{Advances in Neural Information Processing Systems},
  35:32353--32368.

\bibitem[{Wu et~al.(2023)Wu, Manning, and Potts}]{wu-etal-2023-recogs}
Zhengxuan Wu, Christopher~D. Manning, and Christopher Potts. 2023.
\newblock \href {https://doi.org/10.1162/tacl_a_00623} {{R}e{COGS}: How
  incidental details of a logical form overshadow an evaluation of semantic
  interpretation}.
\newblock \emph{Transactions of the Association for Computational Linguistics},
  11:1719--1733.

\bibitem[{Yih et~al.(2014)Yih, He, and Meek}]{yih2014semantic}
Wen-tau Yih, Xiaodong He, and Christopher Meek. 2014.
\newblock Semantic parsing for single-relation question answering.
\newblock In \emph{Proceedings of the 52nd Annual Meeting of the Association
  for Computational Linguistics (Volume 2: Short Papers)}, pages 643--648.

\bibitem[{Yoshikawa et~al.(2017)Yoshikawa, Noji, and
  Matsumoto}]{yoshikawa-etal-2017-ccg}
Masashi Yoshikawa, Hiroshi Noji, and Yuji Matsumoto. 2017.
\newblock \href {https://doi.org/10.18653/v1/P17-1026} {{A}* {CCG} parsing with
  a supertag and dependency factored model}.
\newblock In \emph{Proceedings of the 55th Annual Meeting of the Association
  for Computational Linguistics (Volume 1: Long Papers)}, pages 277--287,
  Vancouver, Canada. Association for Computational Linguistics.

\end{thebibliography}

\appendix

\section{Prompt Template for In-context Learning}\label{app:icl-prompt}

Below is the exact prompt template used for semantic parsing with ICL (described in Section~\ref{subsection:icl}). Placeholders are enclosed in braces.

\begin{lstlisting}[basicstyle=\ttfamily\footnotesize,
                   breaklines=true,
                   breakatwhitespace=false,
                   columns=fullflexible,
                   keepspaces=true,
                   showstringspaces=false,
                   upquote=true]
System message:
You are a precise semantic parser that maps natural language to a logical formula. Respond with only the formula, no explanations.

User message:
Examples:

text: <exemplar_1_text>
formula: <exemplar_1_formula>

text: <exemplar_2_text>
formula: <exemplar_2_formula>

text: <exemplar_3_text>
formula: <exemplar_3_formula>

text: <exemplar_4_text>
formula: <exemplar_4_formula>

text: <exemplar_5_text>
formula: <exemplar_5_formula>

Guidelines for this task:
Output only the logical formula, no explanations.
Use plain predicate and role names (no leading underscores) exactly as in the examples:
dog(x1), run(e1), in(e1,x3), (subj(e1) = x1), (obj(e1) = x2).
Quantification: exists e x.(...). Conjunction: &. Equality: =.
Negation: use the hyphen '-'.
Multiword predicates are single tokens joined with underscores: in_front_of(e,x).
Variables: entities x1,x2,...; events e1,e2,.... Keep parentheses balanced and whitespace minimal.

Now parse the following text to its logical formula.

text: {<SOURCE_TEXT>}
formula:
\end{lstlisting}

\section{Error Examples by Syntactic Features}
\label{appendix:error-examples}

Table~\ref{tab:nsp-errors} shows error examples from the T5-Base model for each syntactic feature.

\begin{table*}[t]
\centering
\caption{Examples of typical errors in neural semantic parsing categorized by syntactic features.}
\label{tab:nsp-errors}
\begin{tabularx}{\textwidth}{lX}
\toprule
\textbf{Error Type} & \textbf{Example (Gold vs. Predicted)} \\
\midrule
Coordinating Conjunctions (CC) 
& \textbf{Sentence:} A man [$_\textsc{conj}$ and] a woman are sitting comfortably on the bench. (\texttt{sick\_test\_706\_p}) \newline
\textbf{Gold:} {\ttfamily exists e1 e2 x3 x4 x5 x6.(man(x3) \& sit(e1) \& (subj(e1) = x3) \& comfortably(e1) \& bench(x4) \& on(e1,x4) \& woman(x5) \& sit(e2) \& (subj(e2) = x5) \& comfortably(e2) \& bench(x6) \& on(e2,x6))} \newline
\textbf{Predicted:} {\ttfamily exists e1 e2 x3 x4 x5 x6.(man(x3) \& sit(e1) \& (subj(e1) = x3) \& comfortably(e1) \& bench(x4) \& on(e1,x4) \& woman(x5) \& sit(e2) \& (subj(e2) = x5) \& comfortably(e2))} \\
\midrule
Prepositional Phrases (PP) 
& \textbf{Sentence:} There is no dog excitedly playing with water  [$_\textsc{pp}$ in the grass.]. (\texttt{sick\_test\_782\_h}) \newline
\textbf{Gold:} {\ttfamily -exists e1 x2 x3 x4.(dog(x2) \& play(e1) \& (subj(e1) = x2) \& water(x3) \& with(e1,x3) \& grass(x4) \& in(e1,x4) \& excitedly(e1))} \newline
\textbf{Predicted:} {\ttfamily -exists e1 x2 x3 x4.(dog(x2) \& play(e1) \& (subj(e1) = x2) \& water(x3) \& with(e1,x3) \& grass(x4) \& in(e1,x4))} \\
\midrule
Passive Voice (PSS) 
& \textbf{Sentence:} A rock is being [$_\textsc{pss}$ climbed] by a person with a rope, which is pink. (\texttt{sick\_test\_642\_h}) \newline
\textbf{Gold:} {\ttfamily exists e1 e2 x3 x4 x5.(rock(x3) \& climb(e1) \& (obj(e1) = x3) \& person(x4) \& rope(x5) \& pink(x5) \& with(e2,x5) \& (subj(e2) = x4) \& (subj(e1) = x4))} \newline
\textbf{Predicted:} {\ttfamily exists e1 x2 x3 x4.(rock(x2) \& climb(e1) \& (obj(e1) = x2) \& person(x3) \& rope(x4) \& pink(x4) \& with(e1,x4))} \\
\bottomrule
\end{tabularx}
\end{table*}

\section{Examples of Quantifier Count Errors}
\label{app:quantifier-error-examples}

Table~\ref{tab:quantifier-errors} presents examples of Quantifier Count errors.

In \texttt{sick\_test\_230\_h}, two distinct event variables should be bound, but only one is introduced; the playing and waiting events are collapsed into a single event, and the variable \texttt{x3} is bound without appearing in the body.

In \texttt{sick\_test\_2872\_p}, the predicate \texttt{mechanical} should take the same variable as \texttt{bull}, but it incorrectly takes a different variable, leading to a mismatch in argument linkage.

\begin{table*}[t]
\centering
\caption{Examples of Quantifier Count errors in predictions.}
\label{tab:quantifier-errors}
\begin{tabularx}{\textwidth}{lX}
\toprule
\textbf{ID} & \textbf{Content} \\
\midrule
\texttt{sick\_test\_230\_h} &
\textbf{Sentence:} There are no children playing and waiting. \\[0.3em]
& \textbf{Gold:} \texttt{-exists e1 e2 x3.(child(x3) \& play(e1) \& (subj(e1) = x3) \& wait(e2) \& (subj(e2) = x3))} \\[0.3em]
& \textbf{Predicted:} \texttt{-exists e1 x2 x3.(child(x2) \& play(e1) \& (subj(e1) = x2) \& wait(e1))} \\
\midrule
\texttt{sick\_test\_2872\_p} &
\textbf{Sentence:} A man is riding a mechanical bull. \\[0.3em]
& \textbf{Gold:} \texttt{exists e1 x2 x3.(man(x2) \& bull(x3) \& mechanical(x3) \& ride(e1) \& (subj(e1) = x2) \& (obj(e1) = x3))} \\[0.3em]
& \textbf{Predicted:} \texttt{exists e1 x2 x3 x4.(man(x2) \& bull(x3) \& mechanical(x4) \& ride(e1) \& (subj(e1) = x2) \& (obj(e1) = x3))} \\
\bottomrule
\end{tabularx}
\end{table*}
\end{document}